\documentclass{new_tlp}
\usepackage{graphicx}
\usepackage{amsmath,amssymb}
\usepackage{dsfont}
\usepackage{theorem}

\theorembodyfont{\normalfont}
\newtheorem{theorem}{Theorem}
\newtheorem{proposition}{Proposition}
\newtheorem{lemma}{Lemma}
\newcommand{\db}{\mathrm{D\hspace{-0.4pt}B}}

\newcommand{\defined}{\stackrel{\rm def}{=}}

\newcommand{\vect}[1]{ {\rm vec}({#1}) }
\newcommand{\mbf}[1]{ {\mathbf #1} }

\newcommand{\denote}[1]{ [\![ #1 ]\!] }

\title[Theory and Practice of Logic Programming]
           {A Linear Algebraic Approach to Datalog Evaluation}

\author[T. Sato]
         {Taisuke Sato\\
          AI research center  AIST / National Institute of Informatics, Tokyo, JAPAN\\
         \email{satou.taisuke@aist.go.jp}
            }

\jdate{}
\pubyear{}
\pagerange{\pageref{00}--\pageref{00}}
\doi{}
\submitted{}
\revised{}
\accepted{}

\begin{document}
\maketitle

\begin{abstract}
In  this paper,  we propose  a fundamentally  new approach  to Datalog
evaluation.  Given  a linear Datalog  program $\db$ written  using $N$
constants  and binary  predicates, we  first translate  if-and-only-if
completions of clauses in $\db$  into a set $\mbf{E}_q(\db)$ of matrix
equations  with  a  non-linear  operation  where  relations  in  ${\bf
  M}_{\db}$,  the  least  Herbrand  model of  $\db$,  are  encoded  as
adjacency matrices.  We then  translate $\mbf{E}_q(\db)$ into another,
but purely  linear matrix  equations $\tilde{\mbf{E}}_q(\db)$.   It is
proved  that the  least  solution of  $\tilde{\mbf{E}}_q(\db)$ in  the
sense  of  matrix ordering  is  converted  to  the least  solution  of
$\mbf{E}_q(\db)$  and the  latter gives  ${\bf M}_{\db}$  as a  set of
adjacency   matrices.   Hence   computing   the   least  solution   of
$\tilde{\mbf{E}}_q(\db)$  is equivalent  to computing  ${\bf M}_{\db}$
specified by  $\db$.  For a class  of tail recursive programs  and for
some  other types  of programs,  our approach  achieves $O(N^3)$  time
complexity irrespective of  the number of variables in  a clause since
only  matrix  operations  costing  $O(N^3)$  or  less  are  used.

We conducted two  experiments that compute the  least Herbrand models
of linear Datalog programs.   The first experiment computes transitive
closure of artificial data and real  network data taken from the Koblenz
Network Collection.  The second one compared the proposed approach with
the state-of-the-art symbolic systems including two Prolog systems and
two ASP systems, in terms of computation time for a transitive closure
program and  the same  generation program.  In  the experiment,  it is
observed  that our  linear algebraic  approach runs  $10^1 \sim  10^4$
times faster than the symbolic systems when data is not sparse.
To appear in Theory and Practice of Logic Programming (TPLP).
\end{abstract}


%\tableofcontents

\section{Introduction}
Top-down  and  bottom-up  have  been   the  two  major  approaches  in
traditional logic programming. They are of contrasting nature but both
compute the  least model  semantics symbolically.   In this  paper, we
propose  a third  approach, a  fundamentally new  one which  evaluates
logic  programs in  vector spaces  to exploit  the potential  of logic
programming in emerging areas.

Given a  class of Datalog  programs $\db$ written using  $N$ constants
and binary  predicates, we first translate  if-and-only-if completions
of  $\db$ into  a  set  $\mbf{E}_q(\db)$ of  matrix  equations in  the
$N$-dimensional  Euclidean  space  $\mathbb{R}^N$  with  a  non-linear
operation.  We  further translate  $\mbf{E}_q(\db)$ into  another, but
purely linear matrix equations $\tilde{\mbf{E}}_q(\db)$.  It is proved
that the  least solution of  $\tilde{\mbf{E}}_q(\db)$ in the  sense of
matrix ordering\footnote{
Matrices  $\mbf{A}=[a_{ij}]$ and  $\mbf{B} =[b_{ij}]$  are ordered  by
$\mbf{A} \leq \mbf{B}$ such that $\mbf{A} \leq \mbf{B}$ if-and-only-if
$a_{ij} \leq b_{ij}$ for all $i,j$.
} can be converted to the least solution of $\mbf{E}_q(\db)$ and the
latter gives the least Herbrand model ${\bf M}_{\db}$ of $\db$ as a
set of adjacency matrices.  We thus can compute ${\bf M}_{\db}$ by way
of solving $\tilde{\mbf{E}}_q(\db)$ algebraically in the vector space
$\mathbb{R}^N$. We emphasize that our approach is not only new but
time complexity wise compared favorably with or better than
conventional Datalog evaluation methods for many important cases as we
discuss later.

Our  approach is  inspired by  the emergence  of big  knowledge graphs
(KGs) such as YAGO  \cite{Suchanek07}, Freebase \cite{Bollacker08} and
Knowledge  Vault  \cite{Dong14}. A  KG  is  a graph  representing  RDF
triples  of the  form $({\rm  subject}:s,\, {\rm  predicate}:p,\, {\rm
  object}:o)$ and  logically speaking, they  are just a set  of ground
atoms $p(s,o)$ with binary predicates.  So one could say that they are
simple.  However the  point is not their logical  simplicity but their
size;  some contain  tens  of  millions of  data,  i.e. ground  atoms.
Researchers working in  the field of KGs  therefore developed scalable
techniques to  cope with huge  KGs, one of  which is a  latent feature
approach that  translates entities and  predicates in the  domain into
vectors, matrices  and tensors  \cite{Kolda09,Cichocki09} respectively
in  vector  spaces  and  apply matrix  and  tensor  decomposition  for
dimension reduction to realize efficient computation \cite{Nickel15}.

Although KGs are just Datalog programs consisting of ground atoms with
binary  predicates and  as  such it  should be  possible  to apply  a
variety of logical  inference, little attention  seems paid to
logical aspects  of KGs.  Only  simple types of logical  inference are
investigated                           so                          far
\cite{Rocktaschel14,Rocktaschel15b,Krompass14,Yang15}.     Thus    the
objective of this  paper is to introduce a  linear algebraic approach
to  logical inference  in vector  spaces, thereby,  bridging KGs  and
logic programming  in general,  or KGs and  Datalog in  particular. By
doing so, we hope to enrich logical  inference for KGs on one hand and
to realize robust and scalable  inference for logic programming on the
other hand.
 
In what  follows, after  a preliminary section,  we describe,  using a
simple tail recursive Datalog program  $\db_{1}$ as a running example,
how  to convert  it to  a  matrix equation  $\mbf{E}_q(\db_1)$ with  a
non-linear  operation in  Section~\ref{Datalog}.  We  then prove  that
$\mbf{E}_q(\db_1)$ is  solvable by  way of  solving an  isomorphic but
purely      linear     equation      $\tilde{\mbf{E}}_q(\db_1)$     in
Section~\ref{Evaluation}.  We generalize our linear algebraic approach
to a more  general class of Datalog programs  than tail-recursive ones
in Section~\ref{Generalization}.  In Section~\ref{Solving}, we examine
subclasses explicitly solvable  in closed form by  linear algebra.  We
validate  our   approach  empirically   through  two   experiments  in
Section~\ref{Experiments}.    In  Section~\ref{Related},   we  briefly
discuss related work  and remaining problems. Section~\ref{Conclusion}
is conclusion.

We assume the reader is familiar  with basics of logic programming and
linear algebra  including tensors \cite{Kolda09,Cichocki09}.   We also
assume that throughout this paper, our first order language ${\cal L}$
for Datalog  programs, i.e., logic programs  without function symbols,
contains  $N$  constants $\{e_1,\ldots,e_N  \}$  and  only $M$  binary
predicates $\{r_1(\cdot,\cdot),\ldots,r_M(\cdot,\cdot) \}$.

\section{Preliminaries}
\label{Preliminaries}

In  this paper,  vectors  are  always column  vectors  and denoted  by
boldface lower  case letters  like ``$\mbf{a}$''.   Similarly matrices
are  square   and  written  by   boldface  upper  case   letters  like
``$\mbf{A}$''.  In particular $\mbf{I}$ is  an identity matrix.  For a
matrix $\mbf{A}  = [a_{ij}]$, put  $(\mbf{A})_{ij} = a_{ij}$.   We use
$\mbf{A}\otimes\mbf{B}$  for the  Kronecker product  of $\mbf{A}$  and
$\mbf{B}$.    ${\rm   vec}({\mbf   A})=   [{\mbf   a}_1^T,\ldots,{\mbf
    a}_M^T]^T$ is  the vectorization  of a matrix  ${\mbf A}  = [{\mbf
    a}_1,\ldots,{\mbf  a}_M]$.    Note  the   fact  that   $\mbf{Y}  =
\mbf{A}\mbf{X}\mbf{B}$  if-and-only-if   ${\rm  vec}({\mathbf   Y})  =
(\mbf{B}^T\otimes \mbf{A})  {\rm vec}({\mathbf  X})$.  We  use $(\cdot
\bullet \cdot)$  for inner  products.  So $(\mbf{a}\bullet  \mbf{b}) =
\mbf{a}^T\mbf{b}$.   $\mathds{1}$ denotes  a matrix  of all  ones.  We
introduce two operator  norms for matrices, $\|  \mbf{A} \|_{\infty} =
\max_i  \sum_j  |a_{ij}|$  and  $\| \mbf{A}  \|_{1}  =  \max_j  \sum_i
|a_{ij}|$.  $\| \mbf{A}^T  \|_{1}  =  \| \mbf{A}  \|_{\infty}$ holds by
definition.

Next  we review  some  logic programming  terminology and  definitions
\cite{Lloyd93}.  Let $\db$ be a Datalog program in a given first-order
language ${\cal L}$ and $\db^g$, the set of ground instances of clauses in $\db$.
Also let ${\bf  HB}$ be the {\em Herbrand base\/}, i.e.,\  the set of all
ground atoms in ${\cal L}$.   Define a mapping $T_{\db}(\cdot): 2^{\bf
  HB} \rightarrow 2^{\bf HB}$ by

\begin{eqnarray*}
T_{\db}(I) & \defined &
  \{\, a \mid \mbox{there is some}\;
       a \leftarrow b_1 \wedge\cdots\wedge b_k \in \db^g \,(k\geq 0) \\
 & &     \hspace{2em}   \mbox{such that}\; \{b_1,\ldots,b_k \} \subseteq I\, \}
\end{eqnarray*}
and a series  $\{ I^{(n)} \}_{n=0,1,\ldots}$ by  $I^{(0)} = \emptyset,
I^{(n+1)} = T_{\db}(I^{(n)})$.  Then we see $I^{(0)} \subseteq I^{(1)}
\subseteq  \cdots$ and  $I^{\infty} =  \bigcup_{n} I^{(n)}$  gives the
{\em least Herbrand model\/} ${\bf M}_{\db}$ of $\db$, least
in the sense of set  inclusion ordering, which is defined by
${\bf M}_{\db} \models a \;\;\mbox{if-and-only-if}\;\; a \in
I^{\infty}$ for any ground atom $a \in {\bf HB}$.\\

Let us  {\em encode} ${\bf  M}_{\db}$, i.e.  isomorphically  map ${\bf
  M}_{\db}$  while preserving  truth values  into the  $N$-dimensional
Euclidean  space  $\mathbb{R}^N$.  Recall  that  the  domain of  ${\bf
  M}_{\db}$ is a set ${\cal D} = \{e_1,\ldots,e_N \}$ of $N$ constants
and        there        are        $M$        binary        predicates
$\{r_1(\cdot,\cdot),\ldots,r_M(\cdot,\cdot)  \}$  in ${\bf  M}_{\db}$.
We translate each $e_i \,(1\leq i\leq N)$ by one-hot encoding into the
$N$-dimensional column  vector $\mbf{e}_i = (0,\cdots  1 \cdots, 0)^T$
in $\mathbb{R}^N$  which has  $1$ as  the $i$-th  element and  $0$ for
other elements.   The set  ${\cal D}' =  \{ \mbf{e}_1,\ldots,\mbf{e}_N
\}$ forms the standard basis of $\mathbb{R}^N$.

Following vector encoding  of domain entities, we  introduce $N \times
N$ adjacency matrices  $\mbf{R}_m \in \{0,1 \}^{N \times  N}$ to encode
relations $r_m(\cdot,\cdot)$ by

\[
(\mbf{R_m})_{ij} =  (\mbf{e}_i \bullet \mbf{R_m}\mbf{e}_j) =
\begin{cases}
    1  & \quad \text{if}\quad {\bf M}_{\db} \models r_m(e_i,e_j) \\
    0  & \quad \text{o.w.}
\end{cases}
\hspace{2em} (1 \leq i,j \leq N, 1 \leq m \leq M)
\]
We say $\mbf{R_m}$ {\em encode\/}s $r_m(\cdot,\cdot)$ in ${\bf M}_{\db}$
and call $\mbf{R_m}$ a matrix encoding $r_m(\cdot,\cdot)$
or representing $r_m(\cdot,\cdot)$.

Now we introduce  the notation $\denote{ F }$, the  truth value of $F$
in ${\bf M}_{\db}$ expressed in terms of vectors and matrices, for a
limited  class of  logical formulas  $F$ used  as the  clause body  of
Datalog programs.  We assume here that  {\em at most two variables are
existentially quantified\/} in the clause  body so that no tensor of
order $n>2$  is required for  the encoding.  Let $x,y,z$  be variables
ranging    over    ${\cal    D}    =    \{e_1,\ldots,e_N    \}$    and
$\mbf{x},\mbf{y},\mbf{z}$ variables  over the domain  of corresponding
one-hot encoding ${\cal D}'= \{\mbf{e}_1,\ldots,\mbf{e}_N\}$.  We use
a non-linear function $\min_1(x)$ defined by
$\min_1(x) =
  \begin{cases}  1 &  \text{if}\quad x\geq 1 \\
                           x &  \text{o.w.} \\
  \end{cases}
$.

Then $\denote{F}$ is defined for a class of AND/OR formulas which is
computed inductively by:

\begin{eqnarray}
\denote{ r(x,y) } & = & (\mbf{x} \bullet \mbf{R}\mbf{y})
    \quad
    \mbox{where $\mbf{R}$ encodes $r(\cdot,\cdot)$ } \label{denote:atom} \\
%\denote{ \neg F } & = & 1 - \denote{F}          \label{denote:neg} \\
\denote{ F_1 \wedge\cdots\wedge F_K }
   & = & \denote{F_1}\cdots\denote{F_K} \label{denote:and} \\
\denote{ F_1 \vee\cdots\vee F_K }
   & = & {\min}_1(\denote{F_1} +\cdots+ \denote{F_K}) \label{denote:or} \\
\denote{ \exists y\, r_i(x,y) \wedge r_j(y,z) }
   & = & {\min}_1 \Big(
    \sum_{k=1}^N (\mbf{x} \bullet \mbf{R}_i\mbf{e_k})(\mbf{e_k}\bullet \mbf{R}_j\mbf{z}) \Big) 
                    \nonumber \\
   & = & {\min}_1 \Big(
    \mbf{x}^T \mbf{R}_i \big(\sum_{k=1}^N \mbf{e_k}\mbf{e_k}^T \big) \mbf{R}_j\mbf{z} \Big)
                    \nonumber \\
   & = & {\min}_1 \big( (\mbf{x} \bullet \mbf{R}_i\mbf{R}_j\mbf{z}) \big)\;\;
     \big(\mbox{as}\; \sum_{k=1}^N \mbf{e_k}\mbf{e_k}^T = \mbf{I} \big).
                    \label{denote:quant}
\end{eqnarray}
Note  that   here  the  existential  quantification   $\exists  y$  is
translated into  $\sum_{k=1}^N \mbf{e_k}\mbf{e_k}^T$  though it  is an
identity matrix.  Now  it is easy to see $\denote{F}  \in \{0,1\}$ and
${\bf  M}_{\db} \models  F$ if-and-only-if  $\denote{F}=1$ for  closed
$F$\footnote{
Formally this is proved by induction on the structure of $F$.
}.

\section{Datalog programs as non-linear matrix equations}
\label{Datalog}

To convey the essential idea quickly,
we use the following simple right recursive
Datalog program $\db_{1}$ as a running example:
\begin{eqnarray}
r_2(x,z) & \leftarrow & r_1(x,z) \nonumber \\
r_2(x,z) & \leftarrow & r_1(x,y)\wedge r_2(y,z)
    \label{prog:trcl}
\end{eqnarray}
$\db_{1}$  computes  the transitive  closure  $r_2(x,y)$  of a  binary
relation $r_1(x,y)$. 

We show  that we are able  to derive a matrix  equation whose solution
gives  $r_2(x,y)$  given  $r_1(x,y)$.   First recall  that  the  least
Herbrand model ${\bf M}_{\db_1}$  of ${\db_1}$ satisfies the following
logical equivalence (called {\em if-and-only-if completion\/} \cite{Lloyd93}):
\begin{eqnarray}
\forall x,z\,
   \left(r_2(x,z) \Leftrightarrow r_1(x,z) \vee \exists\,y\,(r_1(x,y)\wedge r_2(y,z)) \right)
     \label{iff:trcl}
\end{eqnarray}
We translate this equivalence into an equation for
matrices $\mbf{R}_1,\mbf{R}_2$ encoding $r_1(x,z), r_2(y,z)$
as follows.

\begin{eqnarray*}
\denote{r_2(x,z)}
  & = & \denote{r_1(x,z) \vee \exists\,y\,(r_1(x,y)\wedge r_2(y,z))} \;\;
             \mbox{for}\; \forall x,z \in {\cal D} = \{e_1,\ldots,e_N \} \\
\mbox{if-and-only-if} & &\\
(\mbf{x} \bullet \mbf{R}_2\mbf{z})
  & = & {\min}_1( (\mbf{x} \bullet \mbf{R}_1\mbf{z})
              + {\min}_1( (\mbf{x} \bullet \mbf{R}_1\mbf{R}_2\mbf{z}) ) ) \\
  & = & {\min}_1( (\mbf{x} \bullet \mbf{R}_1\mbf{z})
                    + (\mbf{x} \bullet \mbf{R}_1\mbf{R}_2\mbf{z}) ) \\
  & = & {\min}_1( (\mbf{x} \bullet (\mbf{R}_1 + \mbf{R}_1\mbf{R}_2) \mbf{z}) ) \\
  & = & (\mbf{x} \bullet
             {\min}_1( \mbf{R}_1 + \mbf{R}_1\mbf{R}_2 )\mbf{z}) \;\;
                \mbox{for}\; \forall \mbf{x},\mbf{z} \in {\cal  D}'= \{\mbf{e}_1,\ldots,\mbf{e}_N\} \\
\mbox{if-and-only-if} & &\\
\mbf{R}_2
  & = &  {\min}_1( \mbf{R}_1 + \mbf{R}_1\mbf{R}_2 ) \;\;\mbox{for}\;
                          \mbf{R}_1,\mbf{R}_2 \in \{0,1 \}^{N \times N}
\end{eqnarray*}
Here ${\min}_1(\mbf{A})$  for a matrix $\mbf{A}$  means component-wise
application  of ${\min}_1(x)$  function. Note  that ${\min}_1((\mbf{x}
\bullet \mbf{A}\mbf{y})) = (\mbf{x} \bullet {\min}_1(\mbf{A})\mbf{y})$
holds  for any matrix $\mbf{A}$ and  $\mbf{x},\mbf{y} \in
{\cal  D}'  =  \{\mbf{e}_1,\ldots,\mbf{e}_N  \}$.   We  conclude  that
${\mathbf R}_1,\,{\mathbf  R}_2 \in  \{0,1 \}^{N \times  N}$, matrices
encoding  relations   $r_1(\cdot,\cdot),\,r_2(\cdot,\cdot)$  in  ${\bf
  M}_{\db_1}$ respectively, satisfy the following equation:
\begin{eqnarray}
 \mbf{R}_2 & = &  {\min}_1( \mbf{R}_1 + \mbf{R}_1\mbf{R}_2 ).
     \label{matrixeq:trcl}
\end{eqnarray}

We    then   ask    the   converse:    given   $\mbf{R}_1$    encoding
$r_1(\cdot,\cdot)$  in ${\bf  M}_{\db_1}$, does  a matrix  $\mbf{R}_2$
satisfying  (\ref{matrixeq:trcl}) encode  $r_2(\cdot,\cdot)$ in  ${\bf
  M}_{\db_1}$\!?   The  converse is  not  necessarily  true; think  of
$\mbf{R}_2  = \mathds{1}$.   However, fortunately  and evidently,  the
least   solution   $\mbf{R}_2^{*}$  of   (\ref{matrixeq:trcl})   gives
$r_2(\cdot,\cdot)$  in  the  least   model  ${\bf  M}_{\db_1}$.   Here
``least'' means the  ordering among matrices defined  by
for $\mbf{A} =[a_{ij}]$ and $\mbf{B} =[b_{ij}]$,
$\mbf{A} \leq
\mbf{B}$ if-and-only-if $a_{ij} \leq  b_{ij}$ for $\forall \,i,j$.  To
obtain  the  least  solution,  we define  a  series  of  monotonically
increasing matrices ($\in \{0,1  \}^{N \times N}$) $\{ \mbf{R}_2^{(k)}
\}_{k=0,1\ldots}$.

\begin{eqnarray}
\mbf{R}_2^{(0)} & = & \mbf{0}\;\; (\mbox{matrix with every element being $0$})
            \nonumber \\
\mbf{R}_2^{ (k+1) } & = & {\min}_1( \mbf{R}_1 + \mbf{R}_1\mbf{R}_2^{ (k) } ).
          \label{series:trcl}
\end{eqnarray}

Note that $\{ \mbf{R}_2^{(n)} \}$ converges at $n \geq N$. 
It is customary to prove that
the limit $\mbf{R}_2^{(\infty)} = \lim_{n \to \infty} \mbf{R}_2^{(n)} \in \{0,1 \}^{N \times N}$
gives the least solution $\mbf{R}_2^{*}$ of (\ref{matrixeq:trcl}).

\section{Evaluation with linear matrix equations}
\label{Evaluation}

The task of computing the  transitive closure of $r_1(\cdot,\cdot)$ is
now reduced  to computing  the least solution  of the  matrix equation
$\mbf{R}_2   =    {\min}_1(   \mbf{R}_1   +    \mbf{R}_1\mbf{R}_2   )$
(\ref{matrixeq:trcl}) which  is solvable by constructing  a series $\{
\mbf{R}_2^{(k)} \}_{k=0,1\ldots}$.  The  problem is that constructing
$\{ \mbf{R}_2^{(k)} \}_{k=0,1\ldots}$ is  essentially nothing but the
naive bottom-up evaluation of $\db_1$.   We see no clear computational
gain in  solving (\ref{matrixeq:trcl})  by way  of (\ref{series:trcl})
compared to direct  bottom-up evaluation.  What is sought  for here is
to  develop  a  better  evaluation method  than  the  naive  bottom-up
evaluation. Consider an alternative equation:
\begin{eqnarray}
 \tilde{\mbf{R}}_2 & = &  \epsilon( \mbf{R}_1 + \mbf{R}_1\tilde{\mbf{R}}_2 )
     \label{matrixeq:trcl:epsilon}
\end{eqnarray}
where  $\epsilon$ is  a small  positive number  such that 
$(\mbf{I} - \epsilon\mbf{R}_1)^{  -1   }$  exists,
for  example   $\epsilon  < \frac{1}{\| \mbf{R}_1 \|_{\infty}}$.
We   prove    that   the    least   solution    of
(\ref{matrixeq:trcl:epsilon})    gives    the   least    solution    of
(\ref{matrixeq:trcl}).      Define     $\{     \tilde{\mbf{R}}_2^{(k)}
\}_{k=0,1\ldots}$ by

\begin{eqnarray}
\tilde{\mbf{R}}_2^{(0)} & = & \mbf{0}        \nonumber \\
\tilde{\mbf{R}}_2^{ (k+1) } & = & \epsilon( \mbf{R}_1 + \mbf{R}_1\tilde{\mbf{R}}_2^{ (k) } )
          \label{series:trcl:epsilon}
\end{eqnarray}

\begin{lemma}\mbox{}\\
Suppose $0< \epsilon \leq \frac{1}{1+\| \mbf{R}_1 \|_{\infty} }$. Then
$\{ \tilde{\mbf{R}}_2^{(k)} \}_{k=0,1\ldots}$ converges and its limit
$\tilde{\mbf{R}}_2^{(\infty)}$ is
the least solution of (\ref{matrixeq:trcl:epsilon}).
\label{lemma:trcl:1}
\end{lemma}
Proof: We first prove 
$\tilde{\mbf{R}}_2^{(k)} \leq \mathds{1}$ for $\forall\; k \in \mathbb{N}$
by mathematical induction. This obviously holds for $k=0$. 
Suppose $\tilde{\mbf{R}}_2^{(k)} \leq \mathds{1}$. Then 
\begin{eqnarray*}
\tilde{\mbf{R}}_2^{ (k+1) }
 &  =   & \epsilon( \mbf{R}_1 + \mbf{R}_1\tilde{\mbf{R}}_2^{ (k) } ) \\
 & \leq & \epsilon( \mbf{R}_1 + \mbf{R}_1\mathds{1} ) \;\;
               \mbox{by the induction hypothesis}  \\
 & \leq & \epsilon(1+ \| \mbf{R}_1 \|_{\infty} )\mathds{1} \\
 & \leq & \mathds{1} \;\;\; \mbox{because}\; \epsilon(1+\| \mbf{R}_1 \|_{\infty}) \leq 1
\end{eqnarray*}
So $\{ \tilde{\mbf{R}}_2^{(k)} \}_{k=0,1\ldots}$
converges, as a series of monotonically increasing matrices with an upper bound,
to $\tilde{\mbf{R}}_2^{(\infty)}$.
Furthermore,

\begin{eqnarray*}
\tilde{\mbf{R}}_2^{(\infty)}
 & = & \lim_{k \to \infty} \tilde{\mbf{R}}_2^{(k+1)} \\
 & = & \lim_{k \to \infty} \epsilon( \mbf{R}_1 + \mbf{R}_1\tilde{\mbf{R}}_2^{ (k) } ) \\
 & = &  \epsilon( \mbf{R}_1 + \mbf{R}_1 \lim_{k \to \infty} \tilde{\mbf{R}}_2^{ (k) } ) \\
 & = &  \epsilon( \mbf{R}_1 + \mbf{R}_1 \tilde{\mbf{R}}_2^{ (\infty) } )
\end{eqnarray*}
Also let  $\mbf{R}_2'$ be an arbitrary solution of (\ref{matrixeq:trcl:epsilon}).
It can be proved that $\tilde{\mbf{R}}_2^{(k)} \leq \mbf{R}_2'$
holds for $\forall k \in \mathbb{N}$ by mathematical induction.
So $\tilde{\mbf{R}}_2^{(\infty)} = \lim_k \tilde{\mbf{R}}_2^{(k)} \leq \mbf{R}_2'$
is the least solution of (\ref{matrixeq:trcl:epsilon}).
\hspace*{4em} \mbox{Q.E.D.}

\begin{lemma}\mbox{}\\
$\big(\mbf{R}_2^{(k)}\big)_{ij} = 1$ \;if-and-only-if\;
    $\big(\tilde{\mbf{R}}_2^{(k)}\big)_{ij} > 0$
for $\forall\; k \in \mathbb{N}, 1 \leq i,j \leq N$.\footnote{
It is also proved that
$\big(\mbf{R}_2^{(k)}\big)_{ij} = 1$ implies
$\big(\tilde{\mbf{R}}_2^{(k)}\big)_{ij} \geq {\epsilon}^k$
for $\forall\; k \in \mathbb{N}, 1 \leq i,j \leq N$.
}
\label{lemma:trcl:2}
\end{lemma}
Proof: We prove by mathematical induction.
When $k=0$, $\mbf{R}_2^{(0)} = \tilde{\mbf{R}}_2^{(0)} = \mbf{0}$ and
both sides are false.
Suppose $\big(\mbf{R}_2^{(k+1)}\big)_{ij} = 1$.
Then $\big({\min}_1( \mbf{R}_1 + \mbf{R}_1\mbf{R}_2^{ (k) } ) \big)_{ij} = 1$
holds, which implies
$(\mbf{R}_1)_{ij} = 1$ or $\big( \mbf{R}_1\mbf{R}_2^{ (k) } \big)_{ij} \geq 1$.
The latter implies that  for some $m\, (1 \leq m \leq N)$,
$( \mbf{R}_1 )_{im}(\mbf{R}_2^{ (k) })_{mj} = 1$,
and hence $( \mbf{R}_1 )_{im} = 1$ and $(\tilde{\mbf{R}}_2^{ (k) })_{mj} > 0$
by the induction hypothesis.
So $\big( \mbf{R}_1\tilde{\mbf{R}}_2^{ (k) } \big)_{ij} > 0$.
By combining this and $(\mbf{R}_1)_{ij} = 1$ disjunctively, we conclude
$(\tilde{\mbf{R}}_2^{ (k+1) })_{ij} = \epsilon( \mbf{R}_1 + \mbf{R}_1 \tilde{\mbf{R}}_2^{ (k) })_{ij}
= \epsilon( (\mbf{R}_1)_{ij} + (\mbf{R}_1\tilde{\mbf{R}}^{ (k) })_{ij} ) > 0$.
The argument goes the other way around, so we are done. \hspace{3em} Q.E.D.

\begin{theorem}\mbox{}\\
Let $\tilde{\mbf{R}}_2^{*}$ be the least solution of the
matrix equation 
$\tilde{\mbf{R}}_2  =  \epsilon( \mbf{R}_1 + \mbf{R}_1\tilde{\mbf{R}}_2 )$
(\ref{matrixeq:trcl:epsilon})
where $\epsilon$ is a positive number satisfying
$0< \epsilon \leq \frac{1}{1+\| \mbf{R}_1 \|_{\infty} }$.
Define $\mbf{R}_2^{*} \in \{0,1\}^{N \times N}$ by

\[
(\mbf{R}_2^{*})_{ij} = 
\begin{cases} 
\quad 1 & \quad \text{if}\; (\tilde{\mbf{R}}_2^{*})_{ij} > 0 \\
\quad 0 & \quad \text{o.w.} 
\end{cases} 
\quad\quad (1 \leq i,j \leq N)
\]
Then  $\mbf{R}_2^{*}$ is  the least  solution of  the matrix  equation
$\mbf{R}_2   =    {\min}_1(   \mbf{R}_1   +    \mbf{R}_1\mbf{R}_2   )$
(\ref{matrixeq:trcl}).   In other  words, $\mbf{R}_2^{*}$  encodes the
transitive closure of $r_1(\cdot,\cdot)$ in ${\bf M}_{\db_1}$.
\label{theorem:trcl}
\end{theorem}
Proof: By Lemma~\ref{lemma:trcl:1},
$\tilde{\mbf{R}}_2^{*} = \tilde{\mbf{R}}_2^{(\infty)} = \lim_k \tilde{\mbf{R}}_2^{(k)}$.
So for any $i,j\;(1\leq i,j \leq N)$,

\begin{eqnarray*}
(\mbf{R}_2^{*})_{ij} = 1 
& \mbox{if-and-only-if} & ( \tilde{\mbf{R}}_2^{(\infty)} )_{ij}> 0 \\
& \mbox{if-and-only-if} & (\tilde{\mbf{R}}_2^{(k)})_{ij} > 0 \;\mbox{for some}\; k \\
& \mbox{if-and-only-if} & ({\mbf{R}}_2^{(k)})_{ij} = 1 \;\mbox{by Lemma~\ref{lemma:trcl:2}} \\
& \mbox{if-and-only-if} & ({\mbf{R}}_2^{(\infty)})_{ij} = 1.
\end{eqnarray*}
Therefore we have $\mbf{R}_2^{*} = {\mbf{R}}_2^{(\infty)}$.
Since ${\mbf{R}}_2^{(\infty)}$ is the least solution of  (\ref{matrixeq:trcl}),
so is $\mbf{R}_2^{*}$. \hspace{4em}Q.E.D.\\

\begin{proposition}\mbox{}\\
Suppose  $0< \epsilon  \leq  \frac{1}{1+\|  \mbf{R}_1 \|_{\infty}  }$.
Compute $(\mbf{I} - \epsilon\mbf{R}_1)^{  -1 } \epsilon\mbf{R}_1$.  It
coincides   with  the   least   solution  $\tilde{\mbf{R}}_2^{*}$   of
(\ref{matrixeq:trcl:epsilon}) and hence its conversion to a 0-1 matrix
$\mbf{R}_2^{*}$  described  in  Theorem \ref{theorem:trcl}  gives  the
transitive closure of $r_1(\cdot,\cdot)$ in ${\bf M}_{\db_1}$.
\label{prop:trcl}
\end{proposition}
Proof: Since $0< \epsilon \leq \frac{1}{1+\| \mbf{R}_1 \|_{\infty} }$,
$\rho(\epsilon\mbf{R}_1     )$,     the     spectral     radius     of
$\epsilon\mbf{R}_1$,   satisfies    $\rho(\epsilon\mbf{R}_1   )   \leq
\epsilon\|  \mbf{R}_1 \|_{\infty}  \leq  1- \epsilon  <  1$. Consequently
$(\mbf{I} - \epsilon\mbf{R}_1)^{ -1 }$  exists and the matrix equation
(\ref{matrixeq:trcl:epsilon})  has  a   unique  solution  $(\mbf{I}  -
\epsilon\mbf{R}_1)^{ -1 } \epsilon\mbf{R}_1$  which must coincide with
another solution $\tilde{\mbf{R}}_2^{*}$.
\hspace*{12em} \mbox{Q.E.D.}\\

The  choice  of  $\epsilon$  is   arbitrary  but  the  largest  value,
$\frac{1}{1+\| \mbf{R}_1  \|_{\infty}}$, would be preferable  from the
viewpoint  of  the  conversion  of  $\tilde{\mbf{R}}_2^{(\infty)}$  to
$\mbf{R}_2^{*}$.  Note that $\|  \mbf{R}_1 \|_{\infty}$ is the maximum
out-degree  of  nodes  in  $\mbf{R}_1$  as a  graph  and  is  possibly
independent of the graph size.

The time complexity of computing $(\mbf{I} - \epsilon\mbf{R}_1)^{ -1 }
\epsilon\mbf{R}_1$ is $O(N^3)$, or less,  theoretically, if we use the
Coppersmith-Winograd   algorithm   \cite{Coppersmith90}  which   gives
$O(N^{2.376})$.   Hence, we  may say  that in  the case  of transitive
closure computation, our matrix  approach which can be $O(N^{2.376})$,
is  comparable with  or  slightly better  than,  say, tabled  top-down
evaluation of $\db_1$ which requires $O(N^3)$\footnote{
Generally tabled top-down evaluation requires $O(N^{v}) $ where $v$ is
the  maximum number  of  variables of  the body  clause  in a  program
\cite{Warren99}. A deeper  analysis of the time  complexity of Datalog
execution for transitive closure  programs is given in \cite{Tekle10}.
Unfortunately it  concentrates on the  case of  the query of  the form
$\mbox{?-}r_2(x,y)$  where either  $x$ or  $y$ is  ground, and  is not
directly applicable to our case where both $x$ and $y$ are variables.
}.

%<================================
We close  this section with  a concrete example of  transitive closure
computation. Suppose our Herbrand model ${\bf M}_{\db_1}$ of ${\db_1}$
has a  domain ${\cal D} =  \{e_1,\ldots,e_4 \}$ of four  constants and
assume  $\{  r_1(e_1,e_2)$,$r_1(e_2,e_3)$,$r_1(e_3,e_1)$,$r_1(e_4,e_1)
\}$  are  true  w.r.t.\  the relation  $r_1(\cdot,\cdot)$.   Then  the
adjacency matrix $\mbf{R}_1$ encoding $r_1(\cdot,\cdot)$ is given by

\[
  \mbf{R}_1 = \left(
    \begin{array}{rrrr}
      0 & 1 & 0 & 0 \\
      0 & 0 & 1 & 0 \\
      1 & 0 & 0 & 0 \\
      1 & 0 & 0 & 0 \\
    \end{array}
  \right)
\]
and we have $\| \mbf{R}_1 \|_{\infty} = \max \{1,1,1,1\} = 1$.
Put $\epsilon = (1+ \| \mbf{R}_1 \|_{\infty})^{-1} = 1/2$.

\begin{eqnarray*}
\tilde{\mbf{R}}_2^{*}
& = & (\mbf{I} - \epsilon\mbf{R}_1)^{  -1 } \epsilon\mbf{R}_1 \\
& = &
 \left(
 \left(
    \begin{array}{rrrr}
      1 & 0 & 0 & 0 \\
      0 & 1 & 0 & 0 \\
      0 & 0 & 1 & 0 \\
      0 & 0 & 0 & 1 \\
    \end{array}
  \right)
  -
  \left(
    \begin{array}{rrrr}
      0   & 1/2 &   0 &   0 \\
      0   &  0  & 1/2 &   0 \\
      1/2 &  0  &   0 &   0  \\
      1/2 &  0  &   0 &   0 \\
    \end{array}
   \right)
\right)^{-1}
  \left(
    \begin{array}{rrrr}
      0   & 1/2 &   0 &   0 \\
      0   &  0  & 1/2 &   0 \\
      1/2 &  0  &   0 &   0 \\
      1/2 &  0  &   0 &   0 \\
    \end{array}
   \right) \\
& = &
  \left(
    \begin{array}{rrrr}
     0.1428 &  0.5714 &  0.2857 & 0.0000 \\
     0.2857 &  0.1428 &  0.5714 & 0.0000 \\
     0.5714 &  0.2857 &  0.1428 & 0.0000 \\
     0.5714 &  0.2857 &  0.1428 & 0.0000
    \end{array}
   \right)
\end{eqnarray*}   
Hence, by thresholding $\tilde{\mbf{R}}_2^{*}$ at $0$, we reach
\begin{eqnarray*}
\mbf{R}_2^{*}
 & = & 
  \left(
    \begin{array}{rrrr}
     1 & 1 & 1 & 0 \\
     1 & 1 & 1 & 0 \\
     1 & 1 & 1 & 0 \\
     1 & 1 & 1 & 0
    \end{array}
   \right)
\end{eqnarray*}
which is the adjacency matrix encoding the transitive closure of $r_1(\cdot,\cdot)$.
%================================>

\section{Generalization}
\label{Generalization}

$\db_1$ is just  one example of Datalog program.  We  here discuss how
far  we  can  generalize  our linear  algebraic  approach  to  Datalog
evaluation.     We    first    generalize    Lemma~\ref{lemma:trcl:1},
Lemma~\ref{lemma:trcl:2}  and Theorem~\ref{theorem:trcl}  for a  class
${\cal C}_{lin}$  of linear Datalog  programs.  A program $\db$  is in
${\cal C}_{lin}$ if
\begin{itemize}
\item $\db$ contains only binary predicates,
\item non-unit clauses do not contain constants
and take the following form:
$r_0(\{x_0,x_n\}) \leftarrow r_1(\{x_0,x_1\})\wedge\cdots\wedge r_n(\{x_{n-1},x_n\})$
such that $x_0$,\ldots,$x_n$ are all different  and $r_i(\{x_{i-1},x_i\})$
represents either $r_i(x_{i-1},x_i)$ or $r_i(x_i,x_{i-1})$ $(0\leq i \leq n)$,
          and
\item $\db$  is {\em linear\/}  in the following  sense.
\end{itemize}

Let $r$ and  $r'$ be predicates appearing in a  program $\db$.  We say
$r$ depends  on $r'$ if  there is a clause  $H \leftarrow W$  in $\db$
such that  $H$ contains  $r$ and $W$  contains $r'$  respectively, and
write  $r  \succeq_{\db} r'$.   Extend  $r  \succeq r'_{\db}$  to  its
transitive closure (but we use the same symbol $\succeq_{\db}$ for the
closure).   Put $[  r  ]  \defined \{  r'  \mid  r \succeq_{\db}  r'\,
\mbox{and}\, r'  \succeq_{\db} r \}$.   $[ r ]$  is a set  of mutually
dependent predicates  and called  a {\em strongly  connected component
  (SCC) \/} of $\db$.  Predicates in  $\db$ are partitioned into a set
of  SCCs  and SCCs  themselves  are  partially  ordered by  a  partial
ordering, {\em SCC ordering of\/} $\db$, $[r] >_{\db} [r']$ defined as
$[r] >_{\db}  [r']$ if-and-only-if $[r] \succeq_{\db}  [r']$ and $[r']
\not\succeq_{\db} [r]$. We call $\db$ {\em linear\/} if no clause body
contains two  predicates in the  same SCC.

%<=======================================
Linear programs are, intuitively, programs consisting
of clauses such that there is at most one
recursive goal in the clause body.
Note that  checking if $\db$ is linear can  be done
mechanically without difficulty
as the main part, the construction of SCCs, is carried out
by Tarjan's  algorithm\cite{Tarjan72} efficiently  in time  linear in
the number of atoms in $\db$.
%=======================================>
In what follows, we only deal with linear Datalog programs in ${\cal C}_{lin}$.  \\

Let $\db$ be  a linear program in ${\cal C}_{lin}$.   Clauses in $\db$
are  partitioned  into  disjoint sets  $\{  \db'_i  \}_{i=1,\ldots,L}$
called {\em layer\/}s such that  $\db = \db'_1 \cup \cdots\cup \db'_L$
and the head predicates of  $\db'_i\,(1\leq i\leq L)$ coincide with an
SCC which  we denote by $\mbox{SCC}_{\db'_i}$.   Furthermore we assume
no  predicate  in $\db'_i$  depends  on  predicates in  higher  layers
$\db'_j\,(i<j)$.   In other  words $\{  \db'_1,\ldots,\db'_L \}$  is a
list of layers topologically sorted in  the ascending order by the SCC
ordering.   So  the  bottom   layer  program  $\db'_1$  contains  only
predicates minimal  in $\succeq_{\db}$ and  is a union of  ground unit
clauses and clauses of the  form $r(x,y) \leftarrow s(x,y)$ or $r(x,y)
\leftarrow s(y,x)$.

Now fix $\db'_i$. $\db'_i$ consists of clauses of the form:
\begin{eqnarray*}
r(x,y) & \leftarrow & A \\
r(x,y) & \leftarrow & B \wedge s(u,v)\wedge C
\end{eqnarray*}
where  $r,  s   \in  \mbox{SCC}_{\db'_i}$.   $A$,  $B$   and  $C$  are
conjunctions, possibly empty, of atoms whose predicates are defined in
lower  layers $\bigcup_{j  <i} \db'_j$.   Put $\db'_{\leq  i} \defined
\bigcup_{j \leq  i} \db'_j$.  We translate  if-and-only completions of
clauses in  $\db'_i$, which  always hold in  the least  Herbrand model
${\bf  M}_{\db'_{\leq i}}$  of  $\db'_{\leq  i}$ \cite{Lloyd93},  into
matrix  equations  just  like  the  case  of  the  transitive  closure
program~(\ref{matrixeq:trcl}).

Let  $\mbox{SCC}_{\db'_i}   =  \{  r_1,\ldots,r_M  \}$   be  the  head
predicates  of ${\db'_i}$  and $\{  \mbf{R}_{1},\ldots,\mbf{R}_{M} \}$
matrices encoding $\{ r_1,\ldots,r_M \}$ in ${\bf M}_{\db'_{\leq i}}$.
Since a  conjunction of  atoms from lower  layers below  ${\db'_i}$ is
translated  into   a  single   matrix  by  multiplying   matrices,  an
if-and-only completion of clauses in $\db'_i$ is translated into a
matrix equation of the form below:
\begin{eqnarray}
\mbf{R}_{h}
  & = & {\min}_1(F_h[\mbf{R}_{1},\ldots,\mbf{R}_{M}]) \nonumber \\
F_h[\mbf{R}_{1},\ldots,\mbf{R}_{M}]
  & = &  \mbf{A}_1+ \mbf{B}_1 \mbf{R}^{\circ}_{j_1} \mbf{C}_1
                     +\cdots+ \mbf{B}_q \mbf{R}^{\circ}_{j_q}\mbf{C}_q
       \label{matrixeq:gen0}
\end{eqnarray}
Here
$\{ \mbf{R}_{h},\mbf{R}_{j_1},\ldots,\mbf{R}_{j_q} \} \subseteq 
\{ \mbf{R}_{1},\ldots,\mbf{R}_{M} \}$
and $\mbf{R}^{\circ}$ is either $\mbf{R}$ or $\mbf{R}^T$.
$\mbf{A}_1$ is an ${N \times N}$ adjacency matrix 
encoding a disjunction of conjunctions,
while $\mbf{B}_1,\ldots,\mbf{B}_q,\mbf{C}_1,\ldots,\mbf{C}_q$
are ${N \times N}$ adjacency matrices encoding purely conjunctions, and
these conjunctions are made out of
predicates in layers below  ${\db'_i}$.
In summary, $\{ \mbf{R}_{1},\ldots,\mbf{R}_{M} \}$ satisfy
a system $\mbf{E}_q(\db'_i)$ of non-linear matrix equations below
\begin{eqnarray}
\mbf{R}_{1} & = & {\min}_1(F_1[\mbf{R}_{1},\ldots,\mbf{R}_{M}])  \nonumber \\
                    & \cdots &  \nonumber \\
\mbf{R}_{M} & = & {\min}_1(F_M[\mbf{R}_{1},\ldots,\mbf{R}_{M}])
\label{matrixeq:gen}
\end{eqnarray}
where each  $F_h[\mbf{R}_{1},\ldots,\mbf{R}_{M}] \,(1 \leq h \leq M)$
takes the form shown in (\ref{matrixeq:gen0}).\\

Then, conversely, consider $\mbf{E}_q(\db'_i)$ (\ref{matrixeq:gen})
as a set of non-linear matrix equations for unknown
$\{ \mbf{R}_{1},\ldots,\mbf{R}_{M} \}$ and try to solve it.
We define sequences of matrices
$\{ \mbf{R}_{1}^{ (k) },\ldots,\mbf{R}_{M}^{ (k) } \}_{k=0,1\ldots}$
corresponding to (\ref{series:trcl}) by

\begin{eqnarray}
{\mbf{R}}_h^{(0)} & = & \mbf{0} \nonumber \\
{\mbf{R}}_h^{ (k+1) } & = & {\min}_1( F_h[\mbf{R}_1^{ (k) },\ldots,{\mbf{R}}_M^{ (k) }] )
          \label{series:gen}
\end{eqnarray}
for $h\,(1\leq h\leq M)$. We state Lemma~\ref{lemma:gen} without proof.

\begin{lemma}\mbox{}\\
$\{ \mbf{R}_{1}^{ (k) },\ldots,\mbf{R}_{M}^{ (k) } \}_{k=0,1\ldots}$
are monotonically increasing sequences of matrices and
converge to the least solution 
$\{ \mbf{R}_{1}^{ (\infty) },\ldots,\mbf{R}_{M}^{ (\infty) } \}$ of
${\mbf{E}}_q(\db'_i)$ (\ref{matrixeq:gen}).
$\{ \mbf{R}_{1}^{ (\infty) },\ldots,\mbf{R}_{M}^{ (\infty) } \}$
encode $\{ r_1,\ldots,r_M\}$ in the least Herbrand model
${\bf M}_{ \db'_{\leq i} }$ of $ \db'_{\leq i}$.
\label{lemma:gen}
\end{lemma}

Next we introduce,  isomorphically to (\ref{matrixeq:gen}),
a system $\tilde{\mbf{E}}_q(\db'_i)$ of linear matrix equations:
\begin{eqnarray}
\tilde{\mbf{R}}_{1} & = & {\epsilon}_1 F_1[\tilde{\mbf{R}}_{1},\ldots,\tilde{\mbf{R}}_{M}]  \nonumber \\
                    & \cdots &  \nonumber \\
\tilde{\mbf{R}}_{M} & = & {\epsilon}_M F_M[\tilde{\mbf{R}}_{1},\ldots,\tilde{\mbf{R}}_{M}]
                                           \label{matrixeq:gen:epsilon}
\end{eqnarray}
where $\epsilon_h$ is a small positive number satisfying
$\epsilon_h F_h[{\mathds{1}},\ldots,\mathds{1}] \leq \mathds{1} \,(1\leq h\leq M)$.

Define $\{ \tilde{\mbf{R}}_{1}^{ (k) },\ldots,\tilde{\mbf{R}}_{M}^{ (k) } \}_{k=0,1\ldots}$,
correspondingly to  (\ref{series:trcl:epsilon}), by
\begin{eqnarray}
\tilde{\mbf{R}}_h^{(0)} & = & \mbf{0}   \nonumber \\
\tilde{\mbf{R}}_h^{ (k+1) }
  & = & \epsilon_h F_h[\tilde{\mbf{R}}_1^{ (k) },\ldots,\tilde{\mbf{R}}_M^{ (k) }]
          \label{series:gen:epsilon}
\end{eqnarray}
for $h\,(1\leq h\leq M)$.
Proving Lemma~\ref{lemma:series:gen:epsilon} is straightforward:

\begin{lemma}\mbox{}\\
$\{ \tilde{\mbf{R}}_{1}^{ (k) },\ldots,\tilde{\mbf{R}}_{M}^{ (k) } \}_{k=0,1\ldots}$
are monotonically increasing sequences of matrices with upper bound $\mathds{1}$
and converge to $\{ \tilde{\mbf{R}}_{1}^{ (\infty) },\ldots,\tilde{\mbf{R}}_{M}^{ (\infty) } \}$
that give the least solution of $\tilde{\mbf{E}}_q(\db'_i)$ (\ref{matrixeq:gen:epsilon}).
\label{lemma:series:gen:epsilon}
\end{lemma}
We can  also prove Lemma~\ref{lemma:series:gen:epsilon2}  by analyzing
the  form   of  the   right   hand   side   of  equation   shown   in
(\ref{matrixeq:gen0}) (proof omitted).

\begin{lemma}\mbox{}\\
$\big(\mbf{R}_h^{(k)}\big)_{ij} = 1$ \;if-and-only-if\;
    $\big(\tilde{\mbf{R}}_h^{(k)}\big)_{ij} > 0$
for $\forall\; k \in \mathbb{N}, 1\leq h \leq M, 1 \leq i,j \leq N$.
\label{lemma:series:gen:epsilon2}
\end{lemma}
Finally  from Lemma~\ref{lemma:series:gen:epsilon2}, we conclude
Theorem~\ref{theorem:gen} that generalizes Theorem~\ref{theorem:trcl}
(proof omitted):

\begin{theorem}\mbox{}\\
Let $\db$ be  a linear program in ${\cal C}_{lin}$
partitioned and topologically sorted in the ascending order
as $\db  =  \db'_1  \cup \cdots\cup  \db'_L$
by the SCC ordering of $\db$.
Also let $\mbf{E}_q(\db'_i)$ 
be a system of  non-linear matrix equations (\ref{matrixeq:gen}) for matrices
encoding head predicates  $\{ r_1,\ldots,r_M \}$ of ${\db'_i}$. We
suppose predicates in layers below $\db'_i$ are already computed.

Choose a positive number $\epsilon_h\, (1 \leq h \leq M)$ so that
$0< \epsilon_h F_h[\mathds{1},\ldots,\mathds{1}] \leq \mathds{1}$
holds in ${\mbf{E}}_q(\db'_i)$.
Let $\tilde{\mbf{E}}_q(\db'_i)$ be
a system of linear matrix equations (\ref{matrixeq:gen:epsilon}) and
$\{ \tilde{\mbf{R}}_h^{*} \}_{h=1,\ldots,M}$ be the least solution of 
(\ref{matrixeq:gen:epsilon}). Define, for $h \,(1 \leq h \leq M)$,
$\mbf{R}_h^{*} \in \{0,1\}^{N \times N}$ by

\[
(\mbf{R}_h^{*})_{ij} = 
\begin{cases} 
\quad 1 & \quad \text{if}\; (\tilde{\mbf{R}}_h^{*})_{ij} > 0 \\
\quad 0 & \quad \text{o.w.} 
\end{cases} 
\quad\quad (1 \leq i,j \leq N)
\]
Then $\{ \mbf{R}_h^{*} \}_{h=1,\ldots,M}$ are the least solution of
${\mbf{E}}_q(\db'_i)$ encoding $\{ r_1,\ldots,r_M \}$ in the least Herbrand
model ${\bf M}_{\db'_{\leq i}}$.
\label{theorem:gen}
\end{theorem}

What  Theorem~\ref{theorem:gen} tells  us is  that we  can evaluate  a
Datalog  program  $\db =  \db'_1  \cup  \cdots\cup \db'_L$  in  ${\cal
  C}_{lin}$     by     computing     the     least     solution     of
$\tilde{\mbf{E}}_q(\db'_i)$   in  turn   for  $i=1,\ldots,L$   and  by
converting   resulting   solution   matrices  to   0-1   matrices   by
thresholding.

\section{Solving a system of linear matrix equations}
\label{Solving}

Let $\db$  be a Datalog program  in ${\cal C}_{lin}$ and  write $\db =
\db'_1 \cup \cdots\cup \db'_L$ as before.  Put $\tilde{\mbf{E}}_q(\db)
\defined     \bigcup_{i=1}^L      \tilde{\mbf{E}}_q(\db'_i)$     where
$\tilde{\mbf{E}}_q(\db'_i)$ is a system of linear matrix equations for
the $i$-th layer program $\db'_i$.  $\tilde{\mbf{E}}_q(\db)$ is called
a system of linear matrix equations for $\db$.

We   here   discuss   how   to   compute   the   least   solution   of
$\tilde{\mbf{E}}_q(\db)$, or equivalently, the  least solution of each
$\tilde{\mbf{E}}_q(\db'_i)$ (\ref{matrixeq:gen:epsilon}):
\begin{eqnarray*}
\tilde{\mbf{R}}_{1}
  & = & {\epsilon}_1 F_1[\tilde{\mbf{R}}_{1},\ldots,\tilde{\mbf{R}}_{M}]  \nonumber \\
  & \cdots &   \nonumber  \\
\tilde{\mbf{R}}_{M}
  & = & {\epsilon}_M F_M[\tilde{\mbf{R}}_{1},\ldots,\tilde{\mbf{R}}_{M}].
\end{eqnarray*}
Here $F_h \,(1\leq h \leq M)$ is written as  (\ref{matrixeq:gen0}):
\begin{eqnarray*}
F_h[\tilde{\mbf{R}}_{1},\ldots,\tilde{\mbf{R}}_{M}]
 & = &
   \mbf{A}_1+\mbf{B}_1 \tilde{\mbf{R}}^{\circ}_{j_1} \mbf{C}_1 
        +\cdots+ \mbf{B}_q \tilde{\mbf{R}}^{\circ}_{j_q}\mbf{C}_q.
\end{eqnarray*}

Solving  $\tilde{\mbf{E}}_q(\db'_i)$  is not  a  simple  task and  the
difficulty varies with the form of $\tilde{\mbf{E}}_q(\db'_i)$.  So we
discuss   three program  classes, i.e.,
{\em  tail recursive class\/}, {\em  transposed class\/}
and  {\em  two-sided class\/}, each generating different
types of $\tilde{\mbf{E}}_q(\db)$.
We  explain them subsequently   using   examples.

\subsection{Tail recursive class}

This  class  is  a  direct generalization  of  the  transitive  closure
program.   A  program $\db  =  \db'_1\cup\cdots\cup  \db'_L \in  {\cal
C}_{lin}$ is tail recursive if  each layer program $\db'_i$ consists
of clauses of the form
\begin{eqnarray}
r(x,z) & \leftarrow & s_1(x,y_1) \wedge\cdots\wedge s_n(y_{n-1},z) \nonumber \\
r(x,z) & \leftarrow & s_1(x,y_1) \wedge\cdots\wedge s_n(y_{n-1},y) \wedge t(y,z)
    \label{prog:gen:trec}
\end{eqnarray}
where $r$  and $t$ are  mutually dependent predicates in  $\db'_i$ and
the $s_i(\cdot,\cdot)$'s  are defined  in layers below  $\db'_i$.  The
translation of if-and-only-if completions of these clauses into matrix
equations yields a system of matrix equations of the following form:

\begin{eqnarray}
\tilde{\mbf{R}}_{h} 
 & = & \epsilon_h (\mbf{A}_1+\mbf{B}_0 \tilde{\mbf{R}}_{h} +
     \mbf{B}_1 \tilde{\mbf{R}}_{j_1} +\cdots+ \mbf{B}_q\tilde{\mbf{R}}_{j_q}).
       \label{matrixeq:trec}
\end{eqnarray}
This   is     uniquely      solvable      if     
$\epsilon_h<\frac{1}{\|\mbf{B}_0\|_{\infty}}$ and the solution

\[
\tilde{\mbf{R}}_{h} = (\mbf{I} - \epsilon_h \mbf{B}_0)^{-1}
  \epsilon_h(\mbf{A}_1+
   \mbf{B}_1 \tilde{\mbf{R}}_{j_1} +\cdots+ \mbf{B}_q\tilde{\mbf{R}}_{j_q})
\]
is   computed  in   $O(N^3)$.   By   substituting  the   solution  for
$\tilde{\mbf{R}}_{h}$  in other  matrix  equations,  we can  eliminate
$\tilde{\mbf{R}}_{h}$  and eventually,  by  repeatedly solving  matrix
equations                 $M$                times                 for
$\tilde{\mbf{R}}_{1},\ldots,\tilde{\mbf{R}}_{M}$,   reach   a   unique
solution,  i.e.\   the  least  solution   of  $\tilde{\mbf{E}}_q(\db)$
(details omitted).

\subsection{Transposed class}
Programs   $\db   \in  {\cal   C}_{lin}$   in   this  class   generate
$\tilde{\mbf{E}}_q(\db)$   comprised  of   matrix  equations   of  the
following form:

\begin{eqnarray}
\tilde{\mbf{R}}_{h}
  & = & {\epsilon}_h(
              \mbf{A}_1+\mbf{B}_1 \tilde{\mbf{R}}^{T}_{j_1}
        +\cdots+ \mbf{B}_q \tilde{\mbf{R}}^{T}_{j_q}).
                                 \label{matrixeq:gen:transpose1}
\end{eqnarray}
An example of this class, $\db_{2}$, is shown below.
\begin{eqnarray}
r_2(x,z) & \leftarrow & r_1(x,z) \nonumber \\
r_2(x,z) & \leftarrow & r_1(x,y)\wedge r_2(z,y)
    \label{prog:2}
\end{eqnarray}
The difference  from the transitive closure  program (\ref{prog:trcl})
is that $r_2(y,z)$ in (\ref{prog:trcl}) is replaced by $r_2(z,y)$.  So
the  arguments   of  $r_2(y,z)$   are  interchanged.  In   such  case,
$\tilde{\mbf{R}}_2^T$,  the   transpose  of   $\tilde{\mbf{R}}_2$  for
$r_2(y,z)$,  gives  a  matrix   encoding  $r_2(z,y)$.   Assuming  that
$\mbf{R}_1 \in \{0,1\}^{N \times N}$ for $r_1(\cdot,\cdot)$ is already
computed, the linear matrix equation for $r_2(y,z)$ becomes:
\begin{eqnarray}
\tilde{\mbf{R}}_2
  & = &  \epsilon( \mbf{R}_1 + \mbf{R}_1\tilde{\mbf{R}}_2^T ).
     \label{matrixeq:trans:1}
\end{eqnarray}
To  ensure  (\ref{matrixeq:trans:1}) has  a  least  solution, we  also
assume        $\epsilon$       satisfies        $\epsilon       \leq
\frac{1}{1+\|\mbf{R}_1\|_{\infty}  }$ so  that $\epsilon(  \mbf{R}_1 +
\mbf{R}_1\mathbf{1}^T ) \leq \mathbf{1}$ holds.

One way to solve (\ref{matrixeq:trans:1}) is to
substitute (\ref{matrixeq:trans:1}) into itself, resulting in

\begin{eqnarray}
\tilde{\mbf{R}}_2
 & = &  \epsilon( \mbf{R}_1 + \mbf{R}_1\tilde{\mbf{R}}_2^T ) \nonumber \\
 & = &  \epsilon( \mbf{R}_1 + \mbf{R}_1
                    (\epsilon( \mbf{R}_1^T + \tilde{\mbf{R}}_2\mbf{R}_1^T )) \nonumber  \\
 & = &  \epsilon( \mbf{R}_1 + \epsilon\mbf{R}_1\mbf{R}_1^T)
            + \epsilon^2 \mbf{R}_1\tilde{\mbf{R}}_2\mbf{R}_1^T.
     \label{matrixeq:trans:2}
\end{eqnarray}
(\ref{matrixeq:trans:2}) is a  case of two-sided class  treated in the
next subsection.

Another, more general, way is to transform (\ref{matrixeq:trans:1}) to
a   system   of   matrix  equations   about   $\{   \tilde{\mbf{R}}_2,
\tilde{\mbf{R}}_3  \}$ without  transposition,  by  introducing a  new
matrix $\tilde{\mbf{R}}_3 \defined \tilde{\mbf{R}}_2^T$.  We obtain
\begin{eqnarray}
\tilde{\mbf{R}}_2 & = &  \epsilon( \mbf{R}_1 + \mbf{R}_1\tilde{\mbf{R}}_3 ) \nonumber \\
\tilde{\mbf{R}}_3 & = &  \epsilon( \mbf{R}_1^T + \tilde{\mbf{R}}_2\mbf{R}_1^T ).
     \label{matrixeq:trans:3}
\end{eqnarray}
Again (\ref{matrixeq:trans:3}) is a  case of two-sided class discussed
in the next subsection.

\subsection{Two-sided class}
This is a more general class  and much more difficult to evaluate than
previous classes.  Programs in this class have a recursive goal in the
clause  body which  is sandwiched  between two  or more  non-recursive
goals. For simplicity, we assume they generate matrix equations of the
following form:

\begin{eqnarray}
\tilde{\mbf{R}}_{h}
 & = & {\epsilon}_h(
   \mbf{A}_1+ \mbf{B}_0 \tilde{\mbf{R}}_{h} \mbf{C}_0 
        + \mbf{B}_1 \tilde{\mbf{R}}_{j_1} \mbf{C}_1 
        +\cdots+ \mbf{B}_q \tilde{\mbf{R}}_{j_q}\mbf{C}_q ).
       \label{matrixeq:two-sided:1}
\end{eqnarray}
A typical example is $\db_3$ below.
\begin{eqnarray}
r_2(x,z) & \leftarrow & r_1(x,z) \nonumber \\
r_2(x,z) & \leftarrow & r_1(x,y)\wedge r_2(y,w) \wedge r_3(w,z)
    \label{prog:two-sided}
\end{eqnarray}
The linear matrix equation computing $r_2(y,z)$ becomes
\begin{eqnarray}
 \tilde{\mbf{R}}_2
 & = &  {\epsilon}_2( \mbf{R}_1 + \mbf{R}_1\tilde{\mbf{R}}_2\mbf{R}_3 ).
     \label{matrixeq:two-sided:2}
\end{eqnarray}
We assume $\mbf{R}_1,  \mbf{R}_3 \in \{0,1\}^{N \times  N}$ are already
computed. 

(\ref{matrixeq:two-sided:2}) is an example of class of matrix equation
called {\em  discrete Sylvester  equation} which has  been extensively
studied      in       the      field      of       control      theory
\cite{Bartels72,Golub79,Jonsson02,Saberi07,Simoncini13}.

A condition  on $\epsilon$ for (\ref{matrixeq:two-sided:2})  to have a
unique  solution is  stated in  the literature  using eigen  values of
$\mbf{R}_1$ and $  \mbf{R}_3$, but  we need a
concrete   criterion   to   decide    $\epsilon$.    So   we   rewrite
(\ref{matrixeq:two-sided:2})   to   an  equivalent   vector   equation
(\ref{veceq:two-sided:1}),       using       the       fact       that
$\vect{\mbf{A}\mbf{X}\mbf{B}}     =     (\mbf{B}^T\otimes     \mbf{A})
\vect{{\mathbf X}}$  holds for  any matrices $\mbf{A}$,  $\mbf{X}$ and
$\mbf{B}$.

\begin{eqnarray}
 \vect{\tilde{\mbf{R}}_2} 
  & = &  {\epsilon}_2( \vect{\mbf{R}_1}
          + (\mbf{R}_3^T \otimes \mbf{R}_1 )\vect{\tilde{\mbf{R}}_2} )
     \label{veceq:two-sided:1}
\end{eqnarray}
It   is    now   apparent   that    (\ref{veceq:two-sided:1}),   hence
(\ref{matrixeq:two-sided:2}), is uniquely solvable if ${\epsilon}_2 \|
\mbf{R}_3^T   \otimes  \mbf{R}_1   \|_{\infty}  <   1$,  for   example
${\epsilon}_2
\leq\frac{1}{1+\|\mbf{R}_3\|_{1}\|\mbf{R}_1\|_{\infty}}$\footnote{
Recall that ${\epsilon}_2 \| \mbf{R}_3^T  \otimes \mbf{R}_1 \|_{\infty}
\leq {\epsilon}_2 \| \mbf{R}_3^T \|_{\infty}\|  \mbf{R}_1 \|_{\infty}
= {\epsilon}_2 \| \mbf{R}_3  \|_{1}   \|  \mbf{R}_1   \|_{\infty}$.
}and  the   solution  is   $\vect{\tilde{\mbf{R}}_2}  =(\mbf{I}\otimes
\mbf{I}    -     \epsilon_2(\mbf{R}_3^T    \otimes    \mbf{R}_1))^{-1}
{\epsilon}_2\vect{\mbf{R}_1}$.

However,  be warned  that  computing the  solution  this way  requires
$O(N^6)$   time  because   $\mbf{R}_3^T   \otimes   \mbf{R}_1$  is   a
$\{0,1\}^{N^2  \times   N^2}$  matrix.    Fortunately  we   can  solve
(\ref{matrixeq:two-sided:2})  directly  in   $O(N^3)$  as  a  discrete
Sylvester equation \cite{Granat09}, and hence  we can obtain the least
model  of (\ref{prog:two-sided})  in $O(N^3)$,  an order  of magnitude
faster than $O(N^4)$ required by the tabled top-down evaluation method
\cite{Warren99}.

In    general    we    can    always    convert    matrix    equations 
(\ref{matrixeq:two-sided:1})     to      vector     equations     like 
(\ref{veceq:two-sided:1}) and solve them to  obtain the least model of 
the  original  program,  but  this  process  requires  $O(N^6)$  time, 
prohibitively  large in  practice.   So a  more  desirable and  viable 
approach  is  to solve  (\ref{matrixeq:two-sided:1})  as a set of 
discrete Sylvester equations,  which can be done in  $O(N^3)$ for some 
programs      as     we      have     seen. However, when 
(\ref{matrixeq:two-sided:1}) forms  a system of mutually recursive 
discrete Sylvester equations, solving (\ref{matrixeq:two-sided:1})  
remains a  challenging  task, and regrettably,  is  left  for future 
work.

\section{Experiments}
\label{Experiments}

To empirically  validate our  matrix-based method for  Datalog program
evaluation (we hereafter refer to  our matrix-based method as {\em the
Matrix method\/} or just Matrix), we conduct two experiments\footnote{
All  experiments  are carried  out  on  a  PC with  Intel(R)  Core(TM)
i7-3770@3.40GHz CPU, 28GB memory.
}.  The first  experiment measures the computation time  of Matrix for
transitive closure computation to see if it is usable in practice.  We
use artificial data and real data.  The second one compares Matrix and
the  state-of-the-art symbolic  systems including  two Prolog  systems
(B-Prolog \cite{Zhou10}  and XSB  \cite{Swift12}) and two  ASP systems
(DLV  \cite{Alviano10} and  Clingo  \cite{Gebser14}) in  terms of  the
computation  time  required  for   computing  the  transitive  closure
relation and  the same-generation  relation which is  explained later.
We  use artificial  data.  This  experiment revealed  an advantage  of
Matrix in  speed over the compared  systems in the case  of non-sparse
data.

\subsection{Computation time for transitive closure: Matrix vs. Iteration}
\label{subsec:MatIt}
Suppose $\mbf{R}_1$ is an $N \times N$ adjacency matrix encoding a binary
relation  $r_1(x,y)$.   We  denote   by  ${\rm  trcl}(\mbf{R}_1)$  the
adjacency matrix that encodes the transitive closure of $r_1(x,y)$ and
call  it {\em  the transitive  closure matrix  of\/} $\mbf{R}_1$.   We
consider  here  two  linear   algebraic  methods  of  computing  ${\rm
  trcl}(\mbf{R}_1)$\footnote{
All  matrix computation  here  is  done with  GNU  Octave4.0.0
({\tt https://www.gnu.org/software/octave/}).
}.

The first  one, termed the Iteration  method or just Iteration,  is a
base-line   method    which   is   a   faithful    implementation   of
(\ref{series:trcl}).  It computes the least solution of ${\mbf{R}}_2 =
{\min}_1(\mbf{R}_1 + \mbf{R}_1{\mbf{R}}_2)$ by iterating

\begin{eqnarray*}
\mbf{R}_2^{(0)} & = & \mbf{0} \\
\mbf{R}_2^{ (k+1) } & = & {\min}_1( \mbf{R}_1 + \mbf{R}_1\mbf{R}_2^{ (k) } )
\end{eqnarray*}
\noindent
until   convergence   and  returns   the   converged   result  as   ${\rm
trcl}(\mbf{R}_1)$.

The      second      one       is      Matrix      which      computes
$(\mbf{I}-\epsilon\mbf{R}_1)^{-1}   \epsilon\mbf{R}_1$  $(\epsilon   =
\frac{1}{1+\| \mbf{R}_1 \|_{\infty}} )$,  thresholds matrix entries at
0 as described in Theorem~\ref{theorem:trcl} and returns the resulting
matrix as ${\rm trcl}(\mbf{R}_1)$.   $O(N^3)$ time is the theoretically
expected time complexity but may deviate due to implementation details.
We  apply these  two methods  to compute  ${\rm trcl}(\mbf{R}_1)$  and
measure their computation time.

Prior  to the  experiment, we  conducted a  preliminary experiment  to
verify the  correctness  of the  Matrix  method.   We  implemented
Warshall's algorithm as a third method which is a well-known algorithm
for computing  transitive closure  in $O(N^3)$. We applied  all three
methods, i.e.\ Matrix, Iteration  and Warshall's algorithm, to various
$N \times  N$ matrices $\mbf{R}_1$  to see  if they generate  the same
${\rm trcl}(\mbf{R}_1)$.  $\mbf{R}_1$s were randomly generated in such
a way  that for $\forall  i,j (1  \leq i,j \leq  N), \mbf{R}_1(i,j)=1$
with a probability $p_e$ ({\em edge probability\/})\footnote{
$\mbf{R}_1$ encodes  an Erd\H{o}s-R\'enyi  random graph,  a well-known
type of random graphs and it has $p_eN^2$ edges on average.
}.  We  tested various $p_e$  and $N$ up to  $N = 10^3$  and confirmed
that   all   three   methods   agree  and   yield   the   same
${\rm trcl}(\mbf{R}_1)$.

\begin{figure}[h]
\begin{center}
%\begin{minipage}{1.0\hsize}
\hspace*{9em}
\includegraphics[scale=1.0]{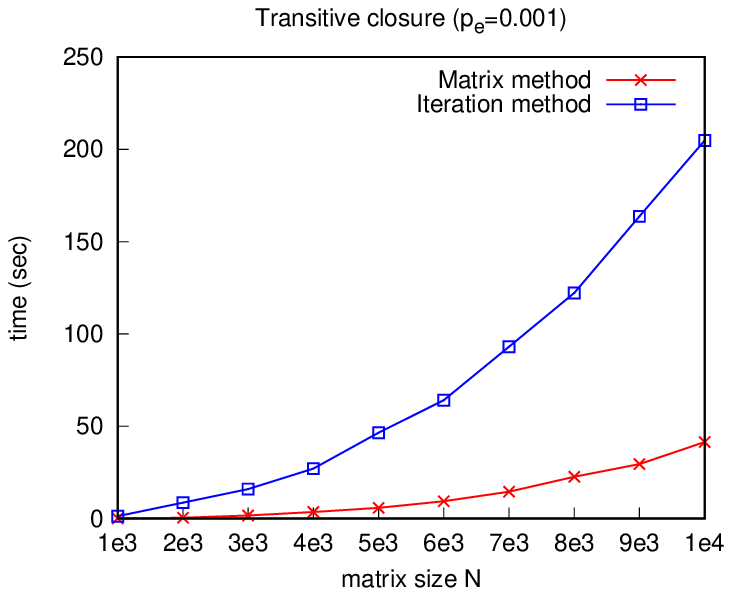}
%\end{minipage}\\
\vspace*{7em}
\caption{Average computation time for ${\rm trcl}(\mbf{R}_1)$}
\label{fig:graph:trcl}
\end{center}
\end{figure}

After having checked the correctness  of Matrix, we compare Matrix and
Iteration.  For each of various $N$s ranging from $10^3$ to $10^4$, we
generate   $\mbf{R}_1$  randomly   with  a   fixed  edge   probability
$p_e=0.001$ and record the computation time for ${\rm trcl}(\mbf{R}_1)$
by  Matrix and  Iteration respectively.   We repeat  this process  five
times and  plot the average  computation time (sec) w.r.t.\  $N$.  The
result is shown in Fig.~\ref{fig:graph:trcl}.

In the graph, the two methods behave similarly w.r.t.~$N$ though it is
not clear whether their behavior is  $O(N^3)$ or not.  We also observe
that  Matrix  constantly  outperforms   Iteration.   For  example,  at
$N=10^4$(1e4) where $\mbf{R}_1$ has  $10^5$ non-zero entries and ${\rm
  trcl}(\mbf{R}_1)$  has $10^8$  non-zero entries  on average,  Matrix
finishes its computation in 40 seconds and runs five times faster than
Iteration.   This graph shows  that Matrix can deal with
$10^4 \times 10^4$ sized or larger matrices\footnote{
Computation time may possibly change depending on the edge probability
$p_e$ which  determines the  density of $\mbf{R}_1$.   Our observation
however suggests that  Matrix's computation time is  not much affected
by $p_e$.
}.\\

We   also   conduct   a   similar  experiment   of   computing   ${\rm
  trcl}(\mbf{R}_1)$  with  real data.   Datasets  are  taken from  the
Koblenz              Network             Collection              ({\tt
  http://konect.uni-koblenz.de/})\cite{Konect13}.    We  choose   five
network graphs with different characters and convert them to adjacency
matrices  $\mbf{R}_1$.   We  then compute  their  transitive  closure
matrices   ${\rm   trcl}(\mbf{R}_1)$    by   Iteration   and   Matrix.
Table~\ref{tab:trclconnect} summarizes  the result.  There $N$  is the
number of entities.   $|\mbf{R}_1|$ is the number  of non-zero entries
of  $N  \times   N$  matrix  $\mbf{R}_1$  and   similarly  for  $|{\rm
  trcl}(\mbf{R}_1)|$.  Matrix and  Iteration indicate their respective
computation  time.   As  with  the case  of  artificial  data,  Matrix
outperforms Iteration in  speed for all datasets, roughly  by an order
of magnitude.

We  emphasize  that although  this  experiment  is a  proof-of-concept
experiment, the  result is encouraging  and suggests the  potential of
our linear algebraic approach.

\begin{table}[h]
\begin{tabular}{rrrrrr}\\\hline
Dataset      &     &     &     & \multicolumn{2}{c}{ Time (sec) } \\ \cline{5-6}
(Koblenz Network Collection)     & $N$  & $|\mbf{R}_1|$ 
            & $|{\rm trcl}(\mbf{R}_1)|$ & Iteration & Matrix \\ \hline\hline
moreno-blogs &  1,224  &  19,025 &    982,061  &      0.7   &    0.1  \\
reactome     &  6,327  & 147,547 &  4,744,333  &    106.0   &   11.3  \\
dblp-cite    & 12,591  &  49,743 &  7,583,575  &  1,238.1   &   86.4  \\
subelj-cora  & 23,166  &  91,500 & 93,584,386  & 17,893.7   &  497.2  \\
ego-twitter  & 23,370  &  33,101 &    353,882  &  5,481.9   &  654.9  \\  \hline
\end{tabular}
\caption{Transitive closure computation for real datasets}
\label{tab:trclconnect}
\end{table}

\subsection{Comparing with the state-of-the-art systems}

We  next compare  our  linear algebraic  approach  with current  major
symbolic approaches, i.e.\  logic programming and ASP.   We select two
state-of-the-art tabled Prolog systems, B-Prolog8.1 and XSB3.6 and two
state-of-the-art   ASP    systems,   DLV(DEC-17-2012    version)   and
Clingo4.5.4. We let them compute  the least Herbrand models of Datalog
programs and  compare computation  time with  computation time  by the
proposed linear algebraic approach  of computing matrices encoding the
models.

We pick  up two linear Datalog  programs in ${\cal C}_{lin}$  shown in
Fig.~\ref{prog:trclsgen}.   They  are  a  transitive  closure  program
(left)  and  a program  for  computing  the same  generation  relation
(right).   We  assume  that  {\tt r1(X,Z)}  and  {\tt  diag(X,Z)}  are
extensional  predicates  defined  by  a  set  of  ground  atoms.   In
particular we assume {\tt diag(X,Z)} represents equality {\tt X=Z} and
the corresponding ground  atoms are of the form  {\tt diag(a,a)}, {\tt
  diag(b,b),...}

% Common code for B-Prolog and XSB and DLV and Clingo
\begin{figure}[h]
\rule{\textwidth}{0.25mm}\\ % [-1em]
\begin{center}
\begin{tabular}{lll}
\qquad {\tt r2(X,Z):- r1(X,Z).}          & \hspace{4em} & {\tt r2(X,W):- diag(X,W).} \\
\qquad {\tt r2(X,Z):- r1(X,Y),r2(Y,Z).}  & \hspace{4em} & {\tt r2(X,W):- r1(X,Y),r2(Y,Z),r1(W,Z).} \\
\end{tabular}
\end{center}
\rule{\textwidth}{0.25mm} \\ %[-1em]
\caption{Tested programs: transitive closure program (left)
             and same\_generation program (right) \label{prog:trclsgen}}
\end{figure}

Both programs in Fig.~\ref{prog:trclsgen}  define {\tt r2(X,Z)} in the
least  Herbrand   model  for  a   given  {\tt  r1(X,Z)}.    They  look
syntactically similar but are substantially different; the left one is
tail-recursive and hence tail-recursive  optimization is possible from
the viewpoint of Prolog  but the right one is not.   Also the left one
defines as  {\tt r2(X,Z)} an  ancestor relation when {\tt  r1(X,Z)} is
interpreted as a parent relation ({\tt X}  is a parent of {\tt Z}) but
the right  one defines the  same generation relation as  {\tt r2(X,Z)}
({\tt X} and {\tt Z} belong to the same generation).

Given these programs, DLV and Clingo automatically compute their least
Herbrand models  by grounding  followed by  search for  stable models.
However  B-Prolog and  XSB  are  designed to  answer  a  query by  SLD
refutation with tabling.  So to let  them compute the least models for
{\tt  r2(X,Z)},  or  to  compute  all solutions  for  the  query  {\tt
  ?-r2(X,Y)},  we drive  them by  a failure  loop below  and ask  {\tt
  ?-top} to measure computation time\footnote{
B-Prolog, XSB and Clingo display computation time when the computation
terminates. We used {\tt -stats}  option to obtain computation time by
DLV.
}.\\

\begin{tabular}{l}
{\tt top:- r2(X,Y),fail.}\\
{\tt top.}\\
\end{tabular}\\

In   the   experiment,   we   first    use   the   left   program   in
Fig.~\ref{prog:trclsgen} and  measure computation time  for transitive
closure computation.  We set N, the number of entities, to 1000.  Then
we choose $p_e$ and randomly generate an $N \times N$ random adjacency
matrix $\mbf{R}_1$  with edge  probability $p_e$.  Finally  we convert
$\mbf{R}_1$  to a  set  of  ground atoms  $\mbox{EDB({\tt  r1})} =  \{
\mbox{{\tt r1($i$,$j$)}} \mid \mbf{R}_1(i,j)=1, 1 \leq i,j \leq N=1000
\}$.

Next we  run four systems, B-Prolog,  XSB, DLV and Clingo,  to measure
their computation time for the transitive closure relation {\tt r2(X,Y)}
of EDB({\tt r1}).   We also compute a transitive  closure matrix ${\rm
  trcl}(\mbf{R}_1)$ encoding  {\tt r2(X,Y)}  by the Matrix  method and
measure  computation time.   We  repeat this  process  five times  and
compute  average  computation  time  for  each  system.   The  average
computation time  for various $p_e$ is  listed in Table~\ref{tab:trcl}
(column names like Matrix, B-Prolog,  XSB, DLV and Clingo indicate the
used system).\\

\begin{table}[h]
\begin{tabular}{rrrrrr}\\\hline
 $p_e$    &  Matrix  & B-Prolog &    XSB  &   DLV   &   Clingo  \\ \hline\hline
0.0001    &  0.096   &   0.000  &   0.000 &   0.000 &     0.000 \\
0.001     &  0.094   &   0.004  &   0.003 &   0.293 &     0.038 \\
0.01      &  0.117   &   2.520  &   1.746 &  10.657 &    14.618 \\
0.1       &  0.105   &  18.382  &  16.296 &  75.544 &   125.993 \\
1.0       &  0.100   & 188.280  & 137.903 & 483.380 & 1,073.301 \\ \hline
\end{tabular}
\caption{Average computation time for transitive closure computation (sec)}
\label{tab:trcl}
\end{table}

As seen from Table~\ref{tab:trcl},  Matrix finishes transitive closure
computation in almost constant time (0.1 second) irrespective of $p_e$
but  symbolic systems  heavily  depend on  $p_e$.   This is  primarily
because  the  average number  of  ground  atoms  in EDB({\tt  r1})  is
proportional  to  $p_e$.  Note  that $p_e=0$  means  all  entries  in
$\mbf{R}_1$ are 0  whereas $p_e=1.0$ means all  entries in $\mbf{R}_1$
are 1.   For $p_e \leq  0.001$, Matrix  runs slower than  the symbolic
systems but  for $p_e \geq  0.01$, it  overwhelms them, runs  $15 \sim
10^4$ times faster.\\

We conduct a similar experiment  with the same generation program with
$N=1000$   while  varying   $p_e$.    Using  the   right  program   in
Fig.~\ref{prog:trclsgen}  and  systematically changing  {\tt  r1(X,Y)}
defined by  EDB({\tt r1}) just as  the case of the  transitive closure
program, we measure average computation time over five runs to compute
{\tt r2(X,Y)} for each of Matrix, B-Prolog, XSB, DLV and Clingo.

For the symbolic systems, all we need to compute {\tt r2(X,Y)} for the
same generation is to replace the  left program with the right program
in Fig.~\ref{prog:trclsgen}.  However Matrix (now  we use it as a term
referring to our linear algebraic approach) needs to compute the least
fixed point of the matrix equation below:
\begin{eqnarray}
 {\mbf{R}}_2
   & = &  {\min}_1( \mbf{I} + \mbf{R}_1{\mbf{R}}_2\mbf{R}^T_1 )  \\
   &   &  (\mbox{$\mbf{I}$ is an identity matrix}) \nonumber
     \label{matrixeq:two-sided:3}
\end{eqnarray}

We therefore first  solve $ \tilde{\mbf{R}}_2 =  {\epsilon}( \mbf{I} +
\mbf{R}_1\tilde{\mbf{R}}_2\mbf{R}^T_1 )$ with  $\epsilon = \frac{1}{1+
  \|\mbf{R}_1\|^2_{\infty}}$, then threshold $\tilde{\mbf{R}}_2$ at 0 to
obtain ${\mbf{R}}_2$.   Although this equation is  not simply solvable
by the inverse matrix  operation,  it  is still  solvable  as a  discrete
Sylvester  equation.   Average computation  time  for  each system  is
summarized   in   Table~\ref{tab:sgen}.     Here   timeout   signifies
computation required more than one hour and was aborted.

\begin{table}[h]
\begin{tabular}{rrrrrr}\\\hline
 $p_e$ &  Matrix &  B-Prolog &  XSB  &  DLV      & Clingo  \\ \hline\hline
0.0001 &  8.475  &  0.001    & 0.001   & 0.003   &  0.002   \\
0.001  &  8.545  &  0.234    & 0.805   & 0.187   &  0.068   \\
0.01   &  10.160 & 1379.090  & timeout & 139.879 &  195.970 \\
0.1    &  10.546 & timeout   & timeout & timeout &  timeout \\\hline
\end{tabular}
\caption{Average computation time for the same generation computation (sec)}
\label{tab:sgen}
\end{table}

Looking  at Table~\ref{tab:sgen},  one  notices the  same tendency  as
Table~\ref{tab:trcl},   i.e.,  Matrix   takes  almost   constant  time
w.r.t.\  $p_e$ while  the  symbolic systems  drastically change  their
computation time  depending on  $p_e$. Also it  is observed  that when
$p_e$ is small ($\leq 0.001$), Matrix's performance is relatively poor
but for  $p_e \geq 0.01$,  it overwhelms  them, ten times  or hundreds
times faster, just as the case of transitive closure computation.

\section{Related work and discussion}
\label{Related}

%<==================
Applying  linear  algebra to  logical  computation  is not  new.   For
example, the SAT  problem is formulated using matrices  and vectors in
\cite{Lin13}.   Concerning  Datalog,  Ceri  \cite{Ceri89} describes  a
bottom-up evaluation  method which is essentially identical to the one
referred  to as  ``Iteration'' in  Subsection~\ref{subsec:MatIt}.  Our
approach is nether bottom-up nor iterative. It abolishes iteration and
replaces it with inverse matrix application.
%==================>
Also there are a couple of  papers concerning KGs that evaluate ground
atoms in a vector space.
Grefenstette  \cite{Grefenstette13}  for example  successfully  embeds
Herbrand models in tensor spaces but the embedding excludes quantified
formulas;  they need  to be  treated separately  by another  framework
which does not accept nested  quantification.  So his formalism is not
applicable  to  our case  that  embeds  Datalog programs  into  vector
spaces, as Datalog programs can have nested existential quantifiers in
their clause bodies.\\

The most  technically relevant work  is RESCAL\cite{Nickel13,Nickel15}
which represents binary relations  $r(x,y)$ by bilinear form $(\mbf{x}
\bullet \mbf{R}\mbf{y})$.   RESCAL is designed to  perform approximate
inference for truth  values of ground atoms  in low-dimensional vector
spaces and exact inference like ours is not treated.  The work done by
Rocktaschel  et  al.~\cite{Rocktaschel15b}  intersects  our  approach.
They  encode  using one-hot  encoding  pairs  of entities  to  vectors
$\mbf{e}$, binary  relations to  vectors $\mbf{r}$, and  represent the
truth  value  as  the   inner  product  $(\mbf{r}  \bullet  \mbf{e})$.
However, unlike us, their encoding  is intended solely for approximate
inference.  No recursion is considered either.\\

There remain numerous  problems to be tackled  for further development
of  our  linear  algebraic  approach.  For  example  extending  binary
predicates to arbitrary predicates is one of them.  Also extending the
class  of  Datalog programs  beyond  linear  ones  is a  big  problem.
Theoretically, there is no difficulty  in dealing with such non-linear
programs in  a vector space.  The  hitch is the difficulty  in solving
derived matrix  equations.  Consider a non-linear  Datalog program for
transitive closure:\\

\begin{tabular}{l}
{\tt r2(X,Z):- r1(X,Z).}\\
{\tt r2(X,Z):- r2(X,Y),r2(Y,Z)}.\\
\end{tabular}\\[1em]

The least  Herbrand model  of the  above program  is straightforwardly
obtained   by  computing   the  least   solution  of   ${\mbf{R}}_2  =
{\min}_1(\mbf{R}_1 +  \mbf{R}_2\mbf{R}_2)$ using the  Iteration method
we described before.   However if one hopes  for efficient computation
along the line of the Matrix method, we need to compute a non-negative
matrix solution $\tilde{\mbf{R}}_2$ of the following matrix equation:

\begin{eqnarray}
\tilde{\mbf{R}}_2 & = & \epsilon(\mbf{R}_1 + \tilde{\mbf{R}_2}\tilde{\mbf{R}_2})
    \label{matrixeq:nonlinear}
\end{eqnarray}

\noindent
using an appropriate $\epsilon$\footnote{
To ensures the existence of the  least non-negative solution of  (\ref{matrixeq:nonlinear})
for $N  \times N$ matrix $\mbf{R}_1$, $\epsilon = \frac{1}{\|\mbf{R}_1\|_{\infty} +N}$
is enough.
}.
Unfortunately, (\ref{matrixeq:nonlinear}) is  a system of multivariate
polynomial equations such that the  number of variables easily goes up
to  $10^4$ or  larger.  Solving  such equations  exactly  is a  highly
technical  problem  in  general   and  no  off-the-shelf  answer  seems
currently available.

Another concern is  negation.  Our approach is  obviously applicable to
Datalog  programs with  stratified  negation, as  negated atoms  $\neg
r(x,y)$ in lower layers are  expressed by $\mathds{1} - \mbf{R}$ where
$\mbf{R}$ is an  adjacency matrix encoding $r(x,y)$.   If programs are
non-stratified  however, the  Matrix method,  originally designed  for
definite clause programs,  needs to be extended in  a fundamental way,
which is an interesting but challenging future work.

Finally although our approach has been successfully applied to domains
with  tens of  thousands  of  entities where  programs  are in  ${\cal
  C}_{lin}$  and self-recursive  as  evidenced by  the experiments  in
Section~\ref{Experiments}, other cases  require specific consideration
and  implementation.  In  particular, mutual  recursive programs  that
have large  SCCs seem difficult  to deal with  when the size  of their
matrix equations get large.

\section{Conclusion}
\label{Conclusion}

We  introduced  an innovative  linear  algebraic  approach to  Datalog
evaluation  for a  class  ${\cal C}_{lin}$  of  Datalog programs  with
binary predicates and linear recursion.   We showed how to translate a
program  $\db  \in  {\cal  C}_{lin}$  to a  system  of  linear  matrix
equations  $\tilde{\mbf{E}}_q(\db)$ and  proved that  thresholding the
solution matrices of $\tilde{\mbf{E}}_q(\db)$ gives adjacency matrices
encoding the  relations in  the least  Herbrand model  ${\bf M}_{\db}$
computed by $\db$.

The  validity of  our  approach is  empirically  verified through  two
experiments.  The  first experiment computed the  least Herbrand model
of a transitive closure program for artificial data and real data.  It
is confirmed that our approach  can efficiently deal with real network
graphs  containing  more  than  $2 \times  10^4$  nodes.   The  second
experiment compared  our approach  with the  state-of-the-art symbolic
systems including two tabled  Prolog systems (B-Prolog8.1 and XSB3.6),
and two modern ASP systems (DLV(DEC-17-2012 version) and Clingo4.5.4).
We measured  average time for  computing the least Herbrand  models of
two Datalog  programs respectively in  the domain of  $10^3$ constants
using $10^3  \times 10^3$  matrices.  It is  observed that  our linear
algebraic  approach  runs  $10^1  \sim 10^4$  times  faster  than  the
symbolic systems when data is not sparse.

\section*{Acknowledgment}
This research  is supported in part by  a project commissioned by  the New
Energy and Industrial Technology Development Organization (NEDO).

\end{document}